\documentclass[runningheads]{llncs}

\usepackage{graphicx}
\usepackage{comment}
\usepackage{amsmath,amssymb}
\usepackage{color}
\usepackage{url}
\usepackage{hyperref}
\usepackage{graphicx}
\renewcommand{\d}[1]{{\mbox{\boldmath$#1$}}}

\newcommand{\m}[1]{\mathsf{#1}}  % safe and LaTeX-native

\renewcommand{\d}[1]{{\mbox{\boldmath$#1$}}}
\clearpage
\begin{document}

%%%%%%%%%%%%%%%%%%%%% Add submission id, track, and title. %%%%%%%%%%%%%%%%%%%%%

% TODO: Please insert your submission number here
%\def\SubNumber{000}

% TODO: Please uncomment the track this paper will be submitted to, comment all other lines
%\def\GCPRTrack{Main Track}
%\def\GCPRTrack{Special Track: Pattern recognition in the life and natural sciences}
%\def\GCPRTrack{Special Track: Photogrammetry and remote sensing}
%\def\GCPRTrack{Young Researcher's Forum}
%\def\GCPRTrack{Fast Review Track}

% TODO: Replace with your title
\title{Confidence-Filtered Relevance (CFR): An Interpretable and Uncertainty-Aware Machine Learning Framework for Naturalness Assessment in Satellite Imagery}
% You can use \thanks for acknowledgment. Do not add any acknowledgment to the draft 
% version that is used for the review process.  
%\title{Title\thanks{XXX}}

% \ifreview
% 	% ANONYMOUS SUBMISSION FOR REVIEW
% 	% DO NOT MODIFY these for the draft version that is used for the review process.
% 	\titlerunning{GCPR 2025 Submission \SubNumber{}. CONFIDENTIAL REVIEW COPY.}
% 	\authorrunning{GCPR 2025 Submission \SubNumber{}. CONFIDENTIAL REVIEW COPY.}
% 	\author{GCPR 2025 - \GCPRTrack{}}
% 	\institute{Paper ID \SubNumber}
% \else
	\titlerunning{CFR for
Naturalness Assessment}

	\author{Ahmed Emam\inst{1}\orcidID{0009-0001-8371-3414} \\\and Ribana Roscher\inst{2}\orcidID{0000-0003-0094-6210}}
	
	\authorrunning{A. Emam et al.}
	
	\institute{University of Bonn, Bonn 53113, Germany\\
	\email{aemam@uni-bonn.de}\\
	\and Forschungszentrum Jülich, 52425 Jülich, Germany\\
	\email{r.roscher}@fz-juelich.de}
%\fi

\maketitle              % typeset the header of the contribution
\begin{abstract}
Protected natural areas play a vital role in ecological balance and ecosystem services. Monitoring these regions at scale using satellite imagery and machine learning is promising, but current methods often lack interpretability and uncertainty-awareness, and do not address how uncertainty affects naturalness assessment. In contrast, we propose Confidence-Filtered Relevance (CFR), a data-centric framework that combines LRP Attention Rollout with Deep Deterministic Uncertainty (DDU) estimation to analyze how model uncertainty influences the interpretability of relevance heatmaps. CFR partitions the dataset into subsets based on uncertainty thresholds, enabling systematic analysis of how uncertainty shapes the explanations of naturalness in satellite imagery. Applied to the AnthroProtect dataset, CFR assigned higher relevance to shrublands, forests, and wetlands, aligning with other research on naturalness assessment. Moreover, our analysis shows that as uncertainty increases, the interpretability of these relevance heatmaps declines and their entropy grows, indicating less selective and more ambiguous attributions. CFR provides a data-centric approach to assess the relevance of patterns to naturalness in satellite imagery based on their associated certainty.

\end{abstract}

\keywords{Explainable Machine Learning (XAI)  \and Uncertainty  Quantification}

\section{Introduction}

Areas with minimal human influence are vital for ecological balance, biodiversity, and ecosystem services such as water regulation and pollination~\cite{sanderson_human_2002,carruthers2025}. These landscapes also offer cultural and educational value, reinforcing the importance of their protection and management~\cite{anderson_role_2020}. Satellite remote sensing enables large-scale monitoring of such regions, offering high-resolution, repeatable observations. However, assessing naturalness from satellite data is challenging, as human impact often occurs gradually, and traditional classification approaches may oversimplify the continuum between natural and modified areas~\cite{stomberg_exploring_2022,emam_leveraging_2024}.

Machine learning (ML), especially CNNs and Vision Transformers (ViTs), has enhanced the modeling of complex spatial patterns from satellite imagery~\cite{ronneberger_u-net_2015,lary_machine_2016}. Despite their power, many ML models lack interpretability and uncertainty estimation. Without these, predictions may rely on spurious correlations and fail to generalize, raising concerns in high-stakes ecological applications~\cite{lapuschkin2019unmasking,pmlr-v70-gal17a}.

We propose \textbf{Confidence-Filtered Relevance (CFR)}, a framework that integrates Layer-wise Relevance Propagation (LRP) Attention Rollout\cite{chefer_transformer_2021} with Deep Deterministic Uncertainty Estimation (DDU) \cite{mukhoti_deep_2023} to stratify model explanations by confidence. Rather than treating all predictions equally, CFR reveals how relevance varies across confidence levels. Applied to the AnthroProtect\cite{stomberg_exploring_2022} dataset, we find that high-confidence predictions emphasize ecologically consistent classes such as \textit{shrublands}, \textit{wetlands}, and \textit{forests}, while low-confidence ones yield diffuse, less meaningful attributions—demonstrating CFR’s uncertainty-aware interpretable naturalness assessment.

\section{Related work}
\subsection{Naturalness Assessment}

Naturalness is typically quantified by identifying areas with minimal human impact. The Human Influence Index (HII)~\cite{sanderson_human_2002} remains a foundational metric, combining indicators like population and infrastructure at 1~km\textsuperscript{2} resolution. However, its static and coarse nature limits its use in dynamic ecosystems. Other indices, such as the Wilderness Quality Index~\cite{WINTER20101624} and the Naturalness Index (NI)~\cite{ekim_naturalness_2021}, improve spatial resolution but rely on fixed assumptions (e.g., low naturalness for shrublands), which can misrepresent ecological value.

Recent ML-based approaches learn spatial patterns directly from imagery~\cite{stomberg_exploring_2022,emam_leveraging_2024} but often lack interpretability and uncertainty estimates. ~\cite{emam_confident_2024} introduced the Confident Naturalness Explanation (CNE), filtering explanations by model confidence at the feature level. Building on this, we propose \textbf{Confidence-Filtered Relevance (CFR)}, which integrates LRP and Deterministic Uncertainty Estimation (DDU) to provide pixel- and class-level attribution across confidence levels, offering more fine-grained and assumption-free insight into naturalness.

\subsection{Explainability and Uncertainty}

Interpretability is crucial in environmental ML to support transparency and informed decision-making. Explainability techniques (XAI) identify which input features influence model predictions. While model-agnostic methods like LIME~\cite{ribeiro_why_2016} and SHAP~\cite{lundberg_unified_2017} offer flexibility, they struggle with high-dimensional data such as satellite imagery~\cite{adadi_peeking_2018}.

Model-specific methods leverage architectural structure. Grad-CAM~\cite{selvaraju_grad-cam_2017} is effective for CNNs but not for Vision Transformers (ViTs)~\cite{dosovitskiy_image_2021}, which rely on self-attention. We therefore use LRP attention rollout~\cite{chefer_transformer_2021}, which propagates relevance through attention layers to yield class-specific, spatially detailed attribution maps.

Uncertainty estimation further enhances trustworthiness, especially for rare or ambiguous land types. While Bayesian methods are principled, they are often impractical. MC-Dropout~\cite{Gal2015DropoutAA} is simpler but incurs a high inference cost. We adopt Deterministic Uncertainty Estimation (DDU)~\cite{mukhoti_deep_2023}, which estimates epistemic uncertainty from a single forward pass via class-wise feature distance, significantly reducing computational cost compared to sampling-based methods.
Combined with LRP, our CFR framework enables confidence-stratified explanations, improving both robustness and interpretability.

\section{Methodology}
\subsection{Model and Dataset}

\begin{figure}
    \centering
    \includegraphics[width=1\linewidth]{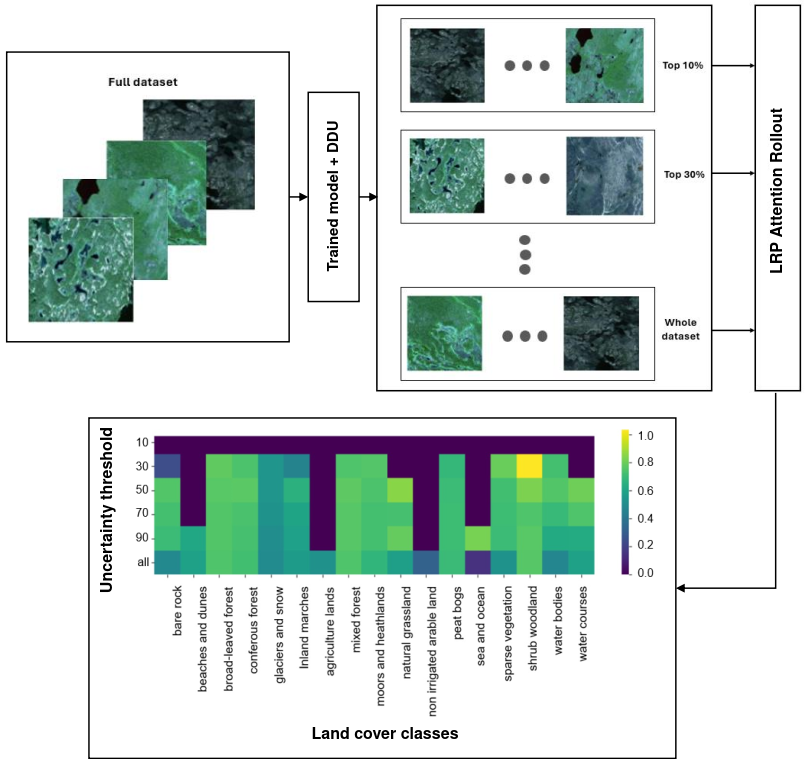}
    \caption{
    \textbf{Overview of the Confidence-Filtered Relevance (CFR) framework.} 
    A Vision Transformer is first trained for naturalness classification, after which DDU is fitted to estimate uncertainty based on the model’s feature space. The dataset is then partitioned into confidence-based subsets, and LRP attention rollout is applied to each. The resulting relevance maps are aggregated by land cover class to analyze how naturalness attribution changes with model confidence.}

    \label{fig:cfr-framework}
\end{figure}

We trained a Vision Transformer (ViT-B/16) using RGB Sentinel-2 patches from the AnthroProtect dataset~\cite{stomberg_exploring_2022}. The model was trained for 50 epochs with a batch size of 64, a learning rate of 2e-5, Adam optimizer, and binary cross-entropy loss. It achieved 99\% test accuracy and an Expected Calibration Error (ECE) of 0.07~\cite{dosovitskiy_image_2021,mukhoti_deep_2023,guo_calibration_2017}. AnthroProtect contains 23{,}919 labeled image patches ($256 \times 256$) from Fennoscandia, with binary \texttt{naturalness} and \texttt{non-naturalness} labels and CORINE-based land cover maps~\cite{european_environment_agency_corine_2019}. We split the dataset into 80\% training, 10\% validation, and 10\% test sets.
Our proposed framework, Confidence-Filtered Relevance (CFR), illustrated in Figure~\ref{fig:cfr-framework}, combines deterministic uncertainty estimation and relevance to naturalness to produce interpretable explanations stratified by model confidence. After training the Vision Transformer, we fit a DDU model~\cite{mukhoti_deep_2023} on its feature representations to estimate uncertainty. We compute a confidence score for each image in the dataset. The dataset is then divided into subsets based on these scores, retaining only the most confident samples within predefined thresholds (e.g., top 10\%, top 30\%). For each subset, we apply LRP attention rollout~\cite{chefer_transformer_2021} to extract class-specific pixel-level relevance maps. Finally, these maps are aggregated over CORINE land cover classes to assess how their contribution to naturalness shifts across different confidence levels.

\subsection{Relevance to Naturalness via LRP Attention Rollout}

To identify class-specific contributions, we employ the attention-based Layer-wise Relevance Propagation (LRP) technique adapted for Vision Transformers by Chefer et al.~\cite{chefer_transformer_2021}. This approach backpropagates relevance from the classification decision through the model’s attention layers. Unlike standard attention rollout~\cite{abnar2020quantifying}, which lacks class sensitivity and often yields ambiguous explanations, this method provides targeted relevance to the \texttt{naturalness} class. 

Let \( A^{(b)} \) denote the attention matrix at transformer block \( b \), and \( R^{(n_b)} \) the relevance at that layer's output. The relevance-adjusted attention is computed as:

\begin{equation}
\bar{\m{A}}^{(b)} = I + \frac{1}{H} \sum_{h=1}^{H} \max\left( \nabla \m{A}^{(b)}_h \odot \m{R}^{(n_b)}_h, 0 \right)
\label{eq:lrp1}
\end{equation}

Equation \ref{eq:lrp1} shows the aggregation of positive relevance-weighted gradients over all attention heads, enhancing interpretability. The identity matrix \(I\) ensures residual connections are preserved, allowing each token to retain part of its own contribution.

The full relevance map is then obtained by recursively combining attention maps from all transformer blocks:
\begin{equation}
    \m{M} = \bar{\m{A}}^{(1)} \cdot \bar{\m{A}}^{(2)} \cdot \ldots \cdot \bar{\m{A}}^{(B)}
\label{eq:lrp22}
\end{equation}

\begin{figure}
    \centering
    \includegraphics[width=\linewidth]{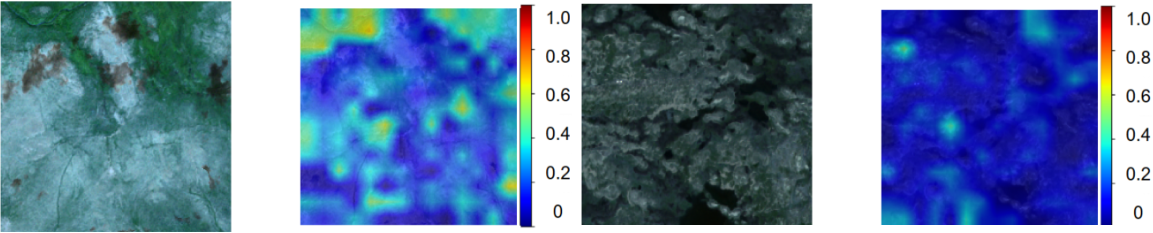}
    \caption{
        \textbf{Visualization of LRP Attention Rollout for the class \texttt{naturalness}.} 
        Columns 1 and 3 show the original Sentinel-2 image patches. 
        Columns 2 and 4 show the corresponding LRP attention rollout maps, 
        highlighting regions with high relevance for the model’s naturalness prediction.
    }
    \label{fig:lrp-rollout}
\end{figure}

Equation~\ref{eq:lrp22} captures how relevance propagates through the model, from output back to input, enabling the identification of regions most responsible for the prediction. Figure \ref{fig:lrp-rollout} shows some image samples and their LRP attention rollout to class naturalness.
\subsection{Uncertainty Estimation via Deep Deterministic Uncertainty}

To quantify uncertainty efficiently, we adopt Deep Deterministic Uncertainty (DDU)~\cite{mukhoti_deep_2023}, a post hoc method that estimates epistemic uncertainty by modeling class-conditional distributions in the feature space. Specifically, we apply DDU to the CLS token embeddings produced by the transformer encoder before classification.

Let $\d{z} \in \mathbb{R}^d$ be the CLS embedding of a sample, and let $\d{\mu}_c \in \mathbb{R}^d$ denote the mean embedding vector for class $c$, computed from the training set. We assume a shared covariance matrix $\m{\Sigma} \in \mathbb{R}^{d \times d}$ across all classes. This corresponds to fitting a multivariate Gaussian distribution to the CLS embeddings for each class. The Mahalanobis distance between $\d{z}$ and class $c$ is given by:

\begin{equation}
    D_c(\d{z}) = \sqrt{(\d{z} - \d{\mu}_c)^{\sf T} \m{\Sigma}^{-1} (\d{z} - \d{\mu}_c)}
    \label{eq:mahalanobis}
\end{equation}

The uncertainty score is then defined as the minimum distance to all class centers:

\begin{equation}
    u(\d{z}) = \min_{c} D_c(\d{z})
    \label{eq:uncertainty}
\end{equation}

Lower values of $u(\d{z})$ indicate embeddings close to a known class distribution and thus higher model confidence, while higher values reflect epistemic uncertainty due to deviation from known training distributions.

We compute the class means $\d{\mu}_c$ and the shared covariance matrix $\m{\Sigma}$ using the training set embeddings. This enables confidence-aware analysis without requiring stochastic sampling or model ensembling. To ensure robustness and semantic consistency, we further enforce Lipschitz continuity in the latent representation of samples classified as \textit{natural}, encouraging similar inputs to yield similar embeddings within the feature space.

\begin{figure}[ht]
    \centering
    \includegraphics[width=\linewidth]{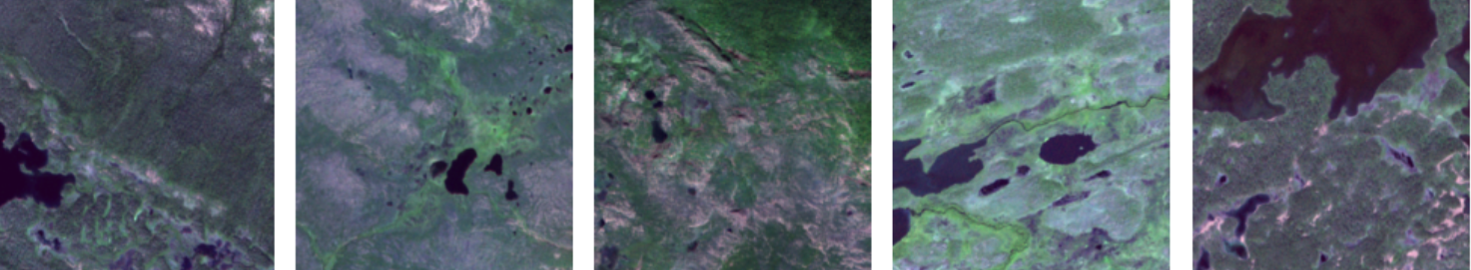} \\[0.5em]
    \includegraphics[width=\linewidth]{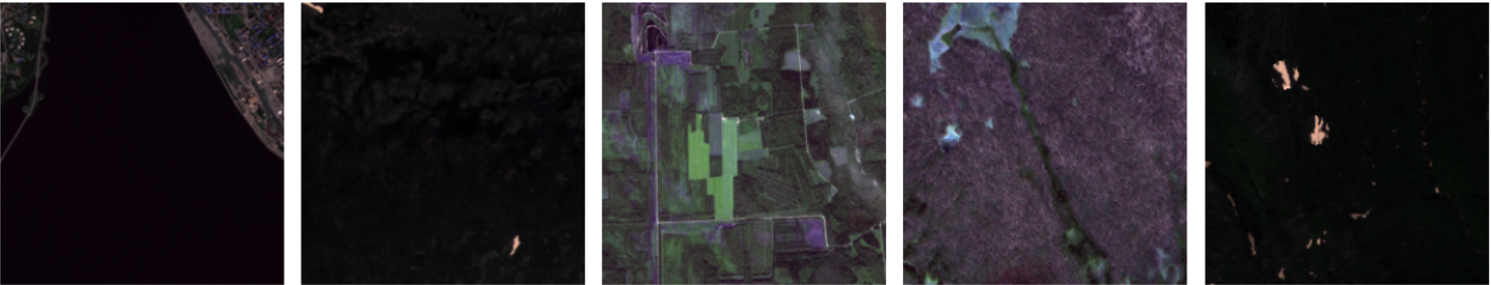}
    \caption{
        \textbf{Examples of model certainty and uncertainty based on DDU scores.} 
        The \textbf{top} subplot shows the images that the model is most certain about, where model embeddings are closest to class means in feature space. 
        These often include cloud-free images with clearly defined vegetative features.
        The \textbf{bottom} subplot shows images associated with the highest uncertainty, typically characterized by clouds, low texture, or visually ambiguous regions such as bare rock or open water, where the model exhibits high epistemic uncertainty due to deviation from known training distributions.
    }
    \label{fig:ddu-certainty}
\end{figure}

Figure~\ref{fig:ddu-certainty} illustrates representative examples from the most certain (top) and most uncertain (bottom) regions based on the DDU uncertainty scores.

\section{Results}

\begin{figure}
    \centering
    \includegraphics[width=1.1\linewidth]{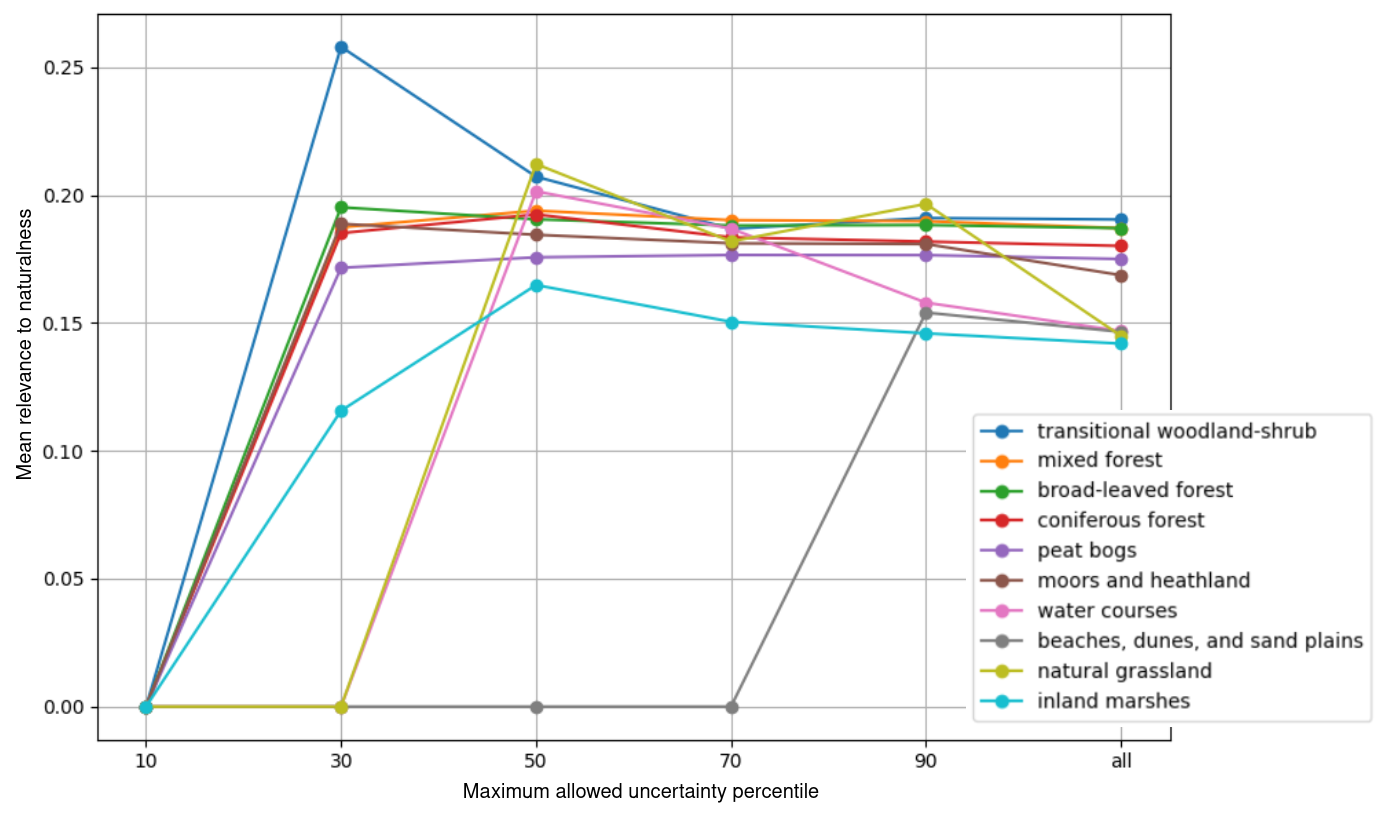} \\[1em]
    \hspace{-0.175\linewidth} % adjust this value as needed
    \includegraphics[width=0.79\linewidth]{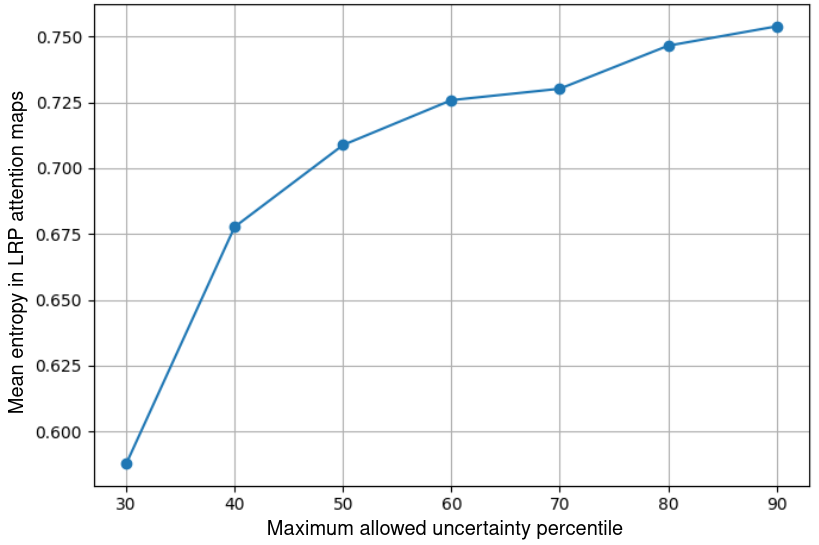}
   \caption{
    \textbf{Influence of Model Confidence on Naturalness Explanations.} 
    \textbf{(Top)} Mean relevance scores for land cover classes across subsets filtered by model uncertainty. Each threshold (e.g., 10\%, 30\%) retains only samples with uncertainty below that percentile, with “all” including the full dataset. High-confidence subsets emphasize ecologically consistent classes like \textit{shrublands} and \textit{wetlands}, while lower-confidence subsets show more dispersed relevance.
    \textbf{(Bottom)} Entropy of the class-level relevance distribution increases with uncertainty, indicating that less confident predictions produce more ambiguous explanations.
    }
    \label{fig:cfr-confidence-figures}
\end{figure}

We evaluate how confidence-filtered explanations shape the relevance to naturalness. Figure~\ref{fig:cfr-confidence-figures} illustrates three perspectives. The \textbf{top plot} shows that high-confidence predictions assign more relevance to ecologically consistent classes like \textit{forests},\texttt{shrubs}, and \texttt{wetlands} while low-confidence predictions yield more ambiguous relevances. The \textbf{bottom plot} quantifies the effect of uncertainty over the relevance entropy, which increases with uncertainty, indicating less selective explanations. Our class uncertainty-aware naturalness relevance rankings correlate strongly with established ecological indicators. We observe a Pearson correlation of 0.91 with the Human Influence Index (HII)~\cite{sanderson_human_2002} using the top 30\% confidence subset, 0.85 with the top 50\%, and only 0.60 without confidence filtering. Similarly, the Confident Naturalness Explanation (CNE) method~\cite{emam_confident_2024}, while not accounting for semantic relationships between land cover classes, still ranks \textit{wetlands}, \textit{shrublands}, and \textit{forests} among the top five contributors, showing strong alignment with our high-confidence results.

\section{Conclusion and Limitations}

Our proposed CFR framework enables interpretable, uncertainty-aware explanations of naturalness from satellite imagery by combining LRP attention rollout with deep deterministic uncertainty estimation. At higher-confidence subsets, CFR consistently highlights ecologically meaningful land cover types such as \textit{shrublands}, \textit{wetlands}, and \textit{forests} as key contributors to naturalness. These findings align with broader ecological research aiming to identify naturalness indicators. Despite its advantages, CFR has several limitations. First, it relies on the assumption that class-conditional embeddings follow a Gaussian distribution, which may not hold in cases of class imbalance or complex feature geometries. Second, CFR estimates only epistemic uncertainty and does not account for aleatoric uncertainty arising from inherent noise or ambiguity in the input data. Future work could enhance the framework by incorporating methods to model both types of uncertainty jointly.

\subsubsection{\ackname}
This work is funded by the Deutsche Forschungsgemeinschaft (DFG, German Research Foundation) RO~4839/5-1 / SCHM~3322/4-1 458156377 - MapinWild, RO 4839/6-1 - 459376902, EXC-2070 - 390732324 - PhenoRob. The authors declare that they have no competing interests.

% Note: if you want to use up all of the allowed space for the paper,
%       the bibliography will start on top of page 13. Furthermore,
%       from page 13 onwards, there will be *only* bibliography, no more
%       figures/tables.
\bibliographystyle{splncs04}
\bibliography{egbib}

\begin{thebibliography}{10}
\providecommand{\url}[1]{\texttt{#1}}
\providecommand{\urlprefix}{URL }
\providecommand{\doi}[1]{https://doi.org/#1}

\bibitem{abnar2020quantifying}
Abnar, S., Zuidema, W.: Quantifying attention flow in transformers. In: Proceedings of the 58th Annual Meeting of the Association for Computational Linguistics (ACL). pp. 4190--4197 (2020)

\bibitem{adadi_peeking_2018}
Adadi, A., Berrada, M.: Peeking {Inside} the {Black}-{Box}: {A} {Survey} on {Explainable} {Artificial} {Intelligence} ({XAI}). IEEE Access  \textbf{6},  52138--52160 (2018). \doi{10.1109/ACCESS.2018.2870052}, \url{https://ieeexplore.ieee.org/document/8466590}, conference Name: IEEE Access

\bibitem{anderson_role_2020}
Anderson, E., Mammides, C.: The role of protected areas in mitigating human impact in the world’s last wilderness areas. Ambio  \textbf{49}(2),  434--441 (Feb 2020). \doi{10.1007/s13280-019-01213-x}, \url{https://doi.org/10.1007/s13280-019-01213-x}

\bibitem{carruthers2025}
Carruthers-Jones, J., et~al.: High-resolution naturalness mapping can support conservation policy objectives across scales. Communications Earth \& Environment  \textbf{6}(279) (2025). \doi{10.1038/s43247-025-02160-0}

\bibitem{chefer_transformer_2021}
Chefer, H., Gur, S., Wolf, L.: Transformer {Interpretability} {Beyond} {Attention} {Visualization}. In: 2021 {IEEE}/{CVF} {Conference} on {Computer} {Vision} and {Pattern} {Recognition} ({CVPR}). pp. 782--791. IEEE, Nashville, TN, USA (Jun 2021). \doi{10.1109/CVPR46437.2021.00084}, \url{https://ieeexplore.ieee.org/document/9577970/}

\bibitem{dosovitskiy_image_2021}
Dosovitskiy, A., Beyer, L., Kolesnikov, A., Weissenborn, D., Zhai, X., Unterthiner, T., Dehghani, M., Minderer, M., Heigold, G., Gelly, S., Uszkoreit, J., Houlsby, N.: An {Image} is {Worth} 16x16 {Words}: {Transformers} for {Image} {Recognition} at {Scale} (Jun 2021). \doi{10.48550/arXiv.2010.11929}, \url{http://arxiv.org/abs/2010.11929}, arXiv:2010.11929

\bibitem{ekim_naturalness_2021}
Ekim, B., Dong, Z., Rashkovetsky, D., Schmitt, M.: The naturalness index for the identification of natural areas on regional scale. International Journal of Applied Earth Observation and Geoinformation  \textbf{105},  102622 (Dec 2021). \doi{10.1016/j.jag.2021.102622}, \url{https://www.sciencedirect.com/science/article/pii/S0303243421003299}

\bibitem{emam_confident_2024}
Emam, A., Farag, M., Roscher, R.: Confident {Naturalness} {Explanation} ({CNE}): {A} {Framework} to {Explain} and {Assess} {Patterns} {Forming} {Naturalness}. IEEE Geoscience and Remote Sensing Letters  \textbf{21} (2024). \doi{10.1109/LGRS.2024.3365196}, \url{https://ieeexplore.ieee.org/document/10433174}, conference Name: IEEE Geoscience and Remote Sensing Letters

\bibitem{emam_leveraging_2024}
Emam, A., Stomberg, T.T., Roscher, R.: Leveraging {Activation} {Maximization} and {Generative} {Adversarial} {Training} to {Recognize} and {Explain} {Patterns} in {Natural} {Areas} in {Satellite} {Imagery}. IEEE Geoscience and Remote Sensing Letters  \textbf{21}, ~1--5 (2024). \doi{10.1109/LGRS.2023.3335473}, \url{https://ieeexplore.ieee.org/document/10325539}

\bibitem{european_environment_agency_corine_2019}
{European Environment Agency}: {CORINE} {Land} {Cover} 2018, {Europe}, 6-yearly - version 2020\_20u1, {May} 2020 (2019). \doi{10.2909/71C95A07-E296-44FC-B22B-415F42ACFDF0}, \url{https://sdi.eea.europa.eu/catalogue/copernicus/api/records/71c95a07-e296-44fc-b22b-415f42acfdf0?language=all}

\bibitem{Gal2015DropoutAA}
Gal, Y., Ghahramani, Z.: Dropout as a bayesian approximation: Representing model uncertainty in deep learning. In: International Conference on Machine Learning (2015)

\bibitem{pmlr-v70-gal17a}
Gal, Y., Islam, R., Ghahramani, Z.: Deep {B}ayesian active learning with image data. In: Precup, D., Teh, Y.W. (eds.) Proceedings of the 34th International Conference on Machine Learning. Proceedings of Machine Learning Research, vol.~70, pp. 1183--1192. PMLR (06--11 Aug 2017), \url{https://proceedings.mlr.press/v70/gal17a.html}

\bibitem{guo_calibration_2017}
Guo, C., Pleiss, G., Sun, Y., Weinberger, K.Q.: On {Calibration} of {Modern} {Neural} {Networks}. CoRR  \textbf{abs/1706.04599} (2017), \url{http://arxiv.org/abs/1706.04599}, arXiv: 1706.04599

\bibitem{lapuschkin2019unmasking}
Lapuschkin, S., W{\"a}ldchen, S., Binder, A., Montavon, G., Samek, W., M{\"u}ller, K.R.: Unmasking clever hans predictors and assessing what machines really learn. Nature Communications  \textbf{10}(1), ~1096 (2019)

\bibitem{lary_machine_2016}
Lary, D.J., Alavi, A.H., Gandomi, A.H., Walker, A.L.: Machine learning in geosciences and remote sensing. Geoscience Frontiers  \textbf{7}(1),  3--10 (Jan 2016). \doi{10.1016/j.gsf.2015.07.003}, \url{https://www.sciencedirect.com/science/article/pii/S1674987115000821}

\bibitem{lundberg_unified_2017}
Lundberg, S.M., Lee, S.I.: A {Unified} {Approach} to {Interpreting} {Model} {Predictions}. In: Advances in {Neural} {Information} {Processing} {Systems}. vol.~30. Curran Associates, Inc. (2017), \url{https://proceedings.neurips.cc/paper_files/paper/2017/hash/8a20a8621978632d76c43dfd28b67767-Abstract.html}

\bibitem{mukhoti_deep_2023}
Mukhoti, J., Kirsch, A., Van~Amersfoort, J., Torr, P.H., Gal, Y.: Deep {Deterministic} {Uncertainty}: {A} {New} {Simple} {Baseline}. In: 2023 {IEEE}/{CVF} {Conference} on {Computer} {Vision} and {Pattern} {Recognition} ({CVPR}). pp. 24384--24394. IEEE, Vancouver, BC, Canada (Jun 2023). \doi{10.1109/CVPR52729.2023.02336}, \url{https://ieeexplore.ieee.org/document/10204383/}

\bibitem{ribeiro_why_2016}
Ribeiro, M., Singh, S., Guestrin, C.: “{Why} {Should} {I} {Trust} {You}?”: {Explaining} the {Predictions} of {Any} {Classifier}. In: DeNero, J., Finlayson, M., Reddy, S. (eds.) Proceedings of the 2016 {Conference} of the {North} {American} {Chapter} of the {Association} for {Computational} {Linguistics}: {Demonstrations}. pp. 97--101. Association for Computational Linguistics, San Diego, California (Jun 2016). \doi{10.18653/v1/N16-3020}, \url{https://aclanthology.org/N16-3020}

\bibitem{ronneberger_u-net_2015}
Ronneberger, O., Fischer, P., Brox, T.: U-{Net}: {Convolutional} {Networks} for {Biomedical} {Image} {Segmentation} (May 2015). \doi{10.48550/arXiv.1505.04597}, \url{http://arxiv.org/abs/1505.04597}

\bibitem{sanderson_human_2002}
Sanderson, E.W., Jaiteh, M., Levy, M.A., Redford, K.H., Wannebo, A.V., Woolmer, G.: The {Human} {Footprint} and the {Last} of the {Wild}: {The} human footprint is a global map of human influence on the land surface, which suggests that human beings are stewards of nature, whether we like it or not. BioScience  \textbf{52}(10),  891--904 (Oct 2002). \doi{10.1641/0006-3568(2002)052[0891:THFATL]2.0.CO;2}, \url{https://doi.org/10.1641/0006-3568(2002)052[0891:THFATL]2.0.CO;2}

\bibitem{selvaraju_grad-cam_2017}
Selvaraju, R.R., Cogswell, M., Das, A., Vedantam, R., Parikh, D., Batra, D.: Grad-{CAM}: {Visual} {Explanations} from {Deep} {Networks} via {Gradient}-{Based} {Localization}. In: 2017 {IEEE} {International} {Conference} on {Computer} {Vision} ({ICCV}). pp. 618--626 (Oct 2017). \doi{10.1109/ICCV.2017.74}, \url{https://ieeexplore.ieee.org/document/8237336}, iSSN: 2380-7504

\bibitem{stomberg_exploring_2022}
Stomberg, T.T., Stone, T., Leonhardt, J., Weber, I., Roscher, R.: Exploring {Wilderness} {Characteristics} {Using} {Explainable} {Machine} {Learning} in {Satellite} {Imagery} (Jul 2022). \doi{10.48550/arXiv.2203.00379}, \url{http://arxiv.org/abs/2203.00379}

\bibitem{WINTER20101624}
Winter, S., Fischer, H.S., Fischer, A.: Relative quantitative reference approach for naturalness assessments of forests. Forest Ecology and Management  \textbf{259}(8),  1624--1632 (2010). \doi{https://doi.org/10.1016/j.foreco.2010.01.040}, \url{https://www.sciencedirect.com/science/article/pii/S0378112710000599}

\end{thebibliography}

\end{document}